\documentclass{article}
\usepackage[utf8]{inputenc}

\usepackage{amsfonts}
\usepackage{amsmath}
\usepackage{amsthm}
\usepackage{float}
\usepackage{graphicx}
\usepackage{subcaption}
\usepackage[linesnumbered,boxed,commentsnumbered,ruled]{algorithm2e}
\usepackage[style=numeric,natbib=true]{biblatex}

\theoremstyle{definition}

\DeclareMathOperator{\kmax}{\text{kmax}}

\renewcommand{\phi}{\varphi}
\renewcommand{\epsilon}{\varepsilon}

\title{Importance Attribution in Neural Networks by Means of Persistence Landscapes of Time Series}

\author{Aina Ferr\`a$\,{}^{1,\,2}$, Carles Casacuberta$\,{}^{1,\,2}$, Oriol Pujol$\,{}^{1,\,2}$}

\date{}

\begin{document}

\footnotetext[1]{Departament de Matem\`atiques i Inform\`atica, Universitat de Barcelona (UB), Gran Via de les Corts Catalanes, 585, 08007 Barcelona, Spain, \{aina.ferra.marcus, carles.casacuberta, oriol\_pujol\} @ub.edu}

\footnotetext[2]{Supported by MCIN/AEI/10.13039/501100011033 under grant PRE2020-094372 (A.\,Ferr\`a) and projects PID2019-105093GB-I00 (A.\,Ferr\`a, O.\,Pujol) and PID2020-117971GB-C22 (C.\,Casacuberta).}
 
\maketitle
\begin{abstract}
We propose and implement a method to analyze time series with a neural network using a matrix 
of area-normalized persistence landscapes obtained through topological data analysis.
We include a gating layer in the network's architecture that is able to identify the most relevant landscape levels for the classification task, thus working as an importance attribution system. Next, we perform a matching between the selected landscape functions and the corresponding critical points of the original time series. From this matching we are able to reconstruct an approximate shape of the time series that gives insight into the classification decision.
We test this technique with input data from a dataset of electrocardiographic signals.
\end{abstract}

\section*{Introduction}

The use of methods from topological data analysis (TDA) to enhance the performance of neural networks is widespread. In some cases the purpose is to provide versatile vectorizations  \cite{perslay}, or to achieve a higher prediction accuracy or classification accuracy  \cite{meryll2019betti}, or
to regularize learning algorithms by feeding topological information extracted from data \cite{chao2019,clough2020,gabrielsson2020}.
Topology has also been used 
to reduce the size of datasets without much loss in training accuracy \cite{rocio2021}. 
A survey of TDA methods for time-series analysis in deep learning using Betti numbers is offered in \cite{umeda2019betti}. 
Tracking changes in the topology of a dataset as it passes through the layers of a well-trained neural network is the subject of \cite{naitzat2020}, while the topology of
neuron activations is analyzed in~\cite{goldfarb2018}. 
Assessment of the generalization gap by means of persistence descriptors without the need of a testing set is discussed in \cite{We,Corneanu2020}.

The use of landscapes as persistence descriptors was initiated by Bubenik in~\cite{bubenik2015landscapes}. Landscapes were used in connection with deep learning in \cite{pllay} with the goal of improving learnability by adding information on topological features of input data into subsequent layers, but not for explainability purposes.
Activation landscapes have also been used as topological summaries of neural network performance in \cite{WBB2021}, and for personalized arrhythmia classification~\cite{yan2019}.

In this article we use TDA towards interpretability of classification results in deep learning. More precisely, we use persistence landscapes to retrieve information about features from data on which a neural network focuses to perform a classification task. 
We preprocess data so that the network is fed with a sequence of landscape levels instead of the original signals. 
The hierarchical structure of persistence landscapes allows us to design a method for finding the most informative levels. For this, we introduce an additional layer to a chosen architecture, whose mission is to assign weights to persistence landscape levels of given signals from a dataset. Then we run again the network using only the landscape levels with the highest weights. The results show that the set of selected landscape levels (normally 2 to 4) yield similar classification accuracies as the collection of the first 10 levels.

Selecting the most relevant landscape levels for a deep learning classification task opens the possibility of reconstructing partially the given data using only the chosen landscape functions. The resulting simplified version of the given data sheds light on which parts of data signals were most relevant for the network's classification task. This provides not only information about the learning process by the network but also about the most essential features carried by data. 

In the context of a heartbeat analysis (Section~\ref{subsection3-2}) we checked that our neural network obtains similar accuracies when fed with reconstructions of signals from selected landscape levels in comparison with those obtained with raw data. This enhances confidence in the classification results by providing evidence that the network is not focusing on 
artifactual details during the learning process.

Our reconstruction method is described more precisely in a companion article \cite{ACO1}, which addresses some mathematical questions related with the present paper and is related with the inverse problem in TDA, namely recovering certain types of data from persistence summaries \cite{,beltonetal,Fasyetal2019,turneretal}.

As an outcome of our approach, we found that the number of landscape levels that are selected as being most relevant for a dataset is inversely associated with the accuracy of a neural network being trained with the given dataset (Fig.\;\ref{fig:relation_weight_accuracy} below). Thus, datasets for which it is unclear which landscape levels should be marked as most important for a deep learning classifier tend to correspond with those with a lower classification performance.

Basic facts about persistence landscapes are collected in Section~\ref{section1}, and our attribution algorithm for landscape levels is described in Section~\ref{section2}.
In Subsection~\ref{subsection3-1} we validate our technique with nine datasets from the UCR Time Series Classification Archive \cite{UCRdataset}, and use it in Subsection~\ref{subsection3-2} to test the accuracy of classification of electrocardiographic signals from the MIT-BIH Arrhytmia Database \cite{MoodyMark2001}. As discussed in
Subsection~\ref{subsection3-3}, shifting signals may cause a loss of classification accuracy by a neural network, while persistence landscapes and results obtained from them remain invariant under horizontal shifts of the data. Hence there is an advantage in using landscapes for classification in cases where such shifts may be due to undesired effects.

\section{Persistence landscapes for sublevel sets}
\label{section1}

Time-series arrays can be viewed as one-dimensional continuous piecewise linear functions where  
persistent homology can be applied to study the evolution of sublevel sets. Thus we consider a sliding parameter $t$ along the $y$-axis, and for each function $f$ defined on an interval $[a,b]$ and each value of $t$ we compute the number of connected components of the corresponding sublevel set $L_t(f)$, which is defined as
\[
L_t(f)=\{x\in [a,b]\mid f(x)\le t\}.
\]
This coincides with the number of connected components of the part of the graph of $f$ which lies at or below height~$t$.
The collection of all sublevel sets for a given function yields a persistence module whose value at $t$ is the vector space $H_0(L_t(f);\mathbb{R})$, where $H_0$ denotes zero-dimensional homology and coefficients in the field $\mathbb{R}$ of real numbers are used.

For background about persistence modules and their associated barcodes and persistence diagrams, see \cite{edelsbrunnerharer2008}.
A \emph{barcode} depicts the lifetime of each connected component of a sublevel set, from the height $t=b$ (birth) where it appears until the height $t=d$ (death) in which it merges with some other connected component. The corresponding \emph{persistence diagram} contains a point $(b,d)$ for each barcode line starting at $b$ and ending at $d$. 
The infinite ray depicting the essential homology class that survives to infinity is discarded for practical purposes.

\begin{figure}[htb]
    \centering
    \begin{subfigure}{.30\textwidth}
        \centering
        \includegraphics[width=\textwidth]{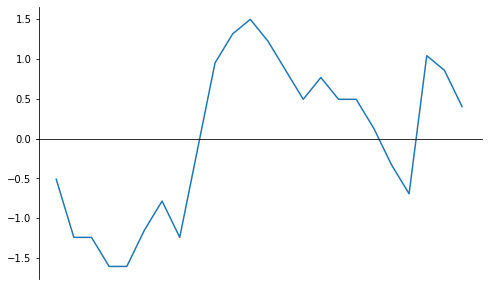}
    \end{subfigure}
    \begin{subfigure}{.38\textwidth}
        \centering
        \includegraphics[width=\textwidth]{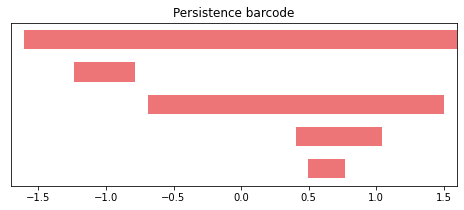}
    \end{subfigure}
    \begin{subfigure}{.30\textwidth}
        \centering
        \includegraphics[width=\textwidth]{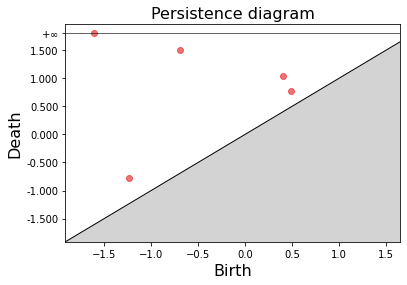}
    \end{subfigure}
    \caption{From left to right, a piecewise linear function, its barcode of zero-dimensional homology of sublevel sets and the corresponding persistence diagram.}
    \label{fig:function_barcode_diagram}
\end{figure}

Barcodes or persistence diagrams are not optimal for their use in deep learning. Neural networks perform best with array-shaped data. 
In this article we use 
\emph{landscapes} as persistence summaries. Persistence landscapes were defined in \cite{bubenik2015landscapes} and, in the case of sublevel sets of signals, they express the evolution of connected components by means of a finite sequence of continuous piecewise linear functions with compact support. Computationally, each landscape function can be expressed as an array of discretized values, which makes it suitable to be introduced into a deep learning system. 

The sequence of landscape functions associated with a persistence diagram is defined as follows. For each point $(b,d)$ in the persistence diagram, one considers the corresponding \emph{tent function}
\[
\Lambda_{(b, d)}(t) = \max \{0, \min \{t-b, d-t\}\}.
\]
Next, a piecewise linear function $\lambda_k \colon\mathbb{R}\to\mathbb{R}$ is defined for each $k\ge 1$ as
\[
\lambda_k(t) = \kmax \{ \Lambda_{(b, d)}(t)\},
\]
where $\kmax$ returns the $k$-th largest value of a given set of real numbers whose elements are counted with multiplicities, or zero if there is no $k$-th largest value. Therefore, since the number of points in a persistence diagram is finite, $\lambda_k=0$ for all sufficiently large values of~$k$.
The first landscape levels $\lambda_1,\lambda_2\dots$ depict the most persistent topological features, while the last ones correspond to less persistent phenomena.

\begin{figure}[htb]
    \centering
    \begin{subfigure}{.24\textwidth}
        \centering
        \includegraphics[width=\textwidth]{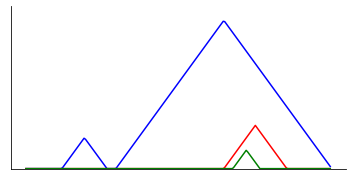}
    \end{subfigure}
    \begin{subfigure}{.24\textwidth}
        \centering
        \includegraphics[width=\textwidth]{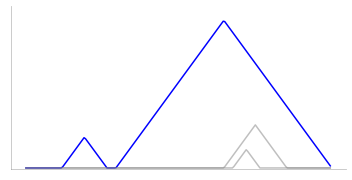}
        
        \vspace{-2cm} $\lambda_1$ \vspace{1.5cm}
                
    \end{subfigure}
    \begin{subfigure}{.24\textwidth}
        \centering
        \includegraphics[width=\textwidth]{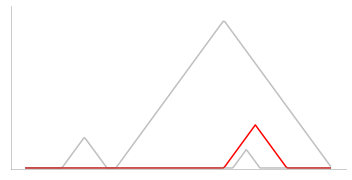}
        
        \vspace{-2cm} $\lambda_2$ \vspace{1.5cm}
                
    \end{subfigure}
    \begin{subfigure}{.24\textwidth}
        \centering
        \includegraphics[width=\textwidth]{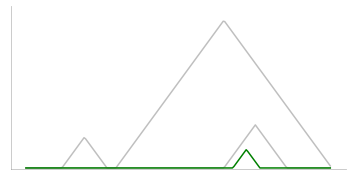}
        
        \vspace{-2cm} $\lambda_3$ \vspace{1.5cm}
        
    \end{subfigure}
    \caption{Sequence of nonzero levels $\lambda_k$ of a persistence landscape (left).}
    \label{fig:sequence}
\end{figure}

\section{Attribution of importance}
\label{section2}

The fact that persistence landscapes can be stratified into a hierarchical sequence of levels makes it possible to design a mechanism for importance attribution ranking landscape levels of a given sample of signals. In \cite{ACO1} a deterministic procedure is described to reconstruct signals from directional persistence landscapes in a number of chosen directions. It is also shown in \cite{ACO1} how to partially reconstruct the given signals using only a subset of selected landscape levels, which is the focus of interest in the present article. By combining this procedure with a machine learning assignment of a sequence of weights to landscapes, we achieve a substantial reduction of the number of critical points of the given data functions without losing much classification accuracy.

To do this, we stack landscape functions from persistence of sublevel sets of the given signals in a matrix that will be fed into a neural network. Landscapes provide a convenient representation, since each landscape level corresponds to a different region of the oscillation of the input signal. 

\begin{figure}[H]
    \centering
    \includegraphics[width=\textwidth]{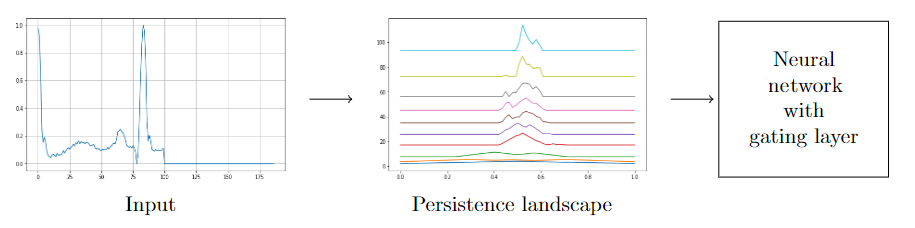}
    \caption{Extracting information through persistence landscapes to feed a neural network.}
    \label{fig:system}
\end{figure}

Since our objective is to feed a deep learning model,  
we decided to normalize the area under each landscape function in order to force the network to focus on their morphology instead of their actual values. This process is illustrated in Fig.\;\ref{fig:system}.

The existence of different levels of information naturally leads to the study of which levels are more important than others for the classification task. In order to implement this idea, we propose the use of a \emph{gating layer}: we maintain the matrix shape throughout all the architecture and, before applying the fully connected layers, each landscape level $\lambda_k$ is multiplied by a positive, less-than-one learnable weight~$w_k$. Thus we obtain a set of weights that indicate how influential is each landscape level for the classification task. Typically, a network should regard the first landscape levels as more important than the last ones, given that the first levels contain information about the most persistent topological features. 

By building a ranking of all the landscape levels, we are able to decide at which threshold of information the network stops learning. This is helpful in two main ways: first, we are able to reduce the information that we use to train our system by reducing the number of landscape functions that we pass to our network; and second, we can attribute importance to the parts of the original data that are producing the most relevant landscape levels.

\section{Experimental setting and results}
\label{section3}

In this section, we present the results of our experiments using a neural network with a fixed architecture and different input signals. Our main aims are to assess the changes in classification accuracy by using only a set of selected landscape levels in comparison with the full landscape and with the original data, while determining which are the most relevant landscape features in each database. Robustness of our method is estimated by applying it to nine databases of very different nature.

\medskip

\noindent
{\bf Data.}
We applied our methodology to a collection of datasets taken from the UCR Time Series Classification Archive \cite{UCRdataset}. The criteria for choosing a dataset 
were the following: the dataset should have at most five different classes and the total number of samples divided by the number of classes should be greater than or equal to~$500$. 
These criteria were adopted in order to avoid dealing with data scarcity problems and difficulties caused by imbalanced classes or by an excessive number of classes. Table \ref{table_dataset} contains a summary of the characteristics of each dataset.

\bgroup
\def\arraystretch{1.1}
\begin{table}[htb]
\centering
\begin{tabular}{l|crcc}
\textbf{Dataset}                      & \textbf{Samples} & \textbf{Length} & \textbf{Classes} & \textbf{Imbalanced} \\ \hline \\[-0.35cm]
\textbf{ECG5000}             & 5000                       & 140\phantom{xi}                        & 5                          & Yes                 \\

\textbf{FreezerRegularTrain} & 3000                       & 301\phantom{xi}                        & 2                          & No                  \\

\textbf{HandOutlines}        & 1370                       & 2709\phantom{xi}                       & 2                          & Yes                 \\

\textbf{ItalyPowerDemand}    & 1096                       & 24\phantom{xi}                         & 2                          & No                  \\

\textbf{MoteStrain}          & 1272                       & 84\phantom{xi}                         & 2                          & No                 \\

\textbf{PhalangesOutlinesCorrect}   & 2658                       & 80\phantom{xi}                         & 2                          & Yes                 \\

\textbf{StarLightCurves}     & 9236                       & 1024\phantom{xi}                       & 3                          & Yes                 \\

\textbf{Wafer}               & 7164                       & 152\phantom{xi}                        & 2                          & Yes                 \\

\textbf{Yoga}                & 3300                       & 426\phantom{xi}                        & 2                          & Yes                \\[0.05cm] \hline 
\end{tabular}
\caption{A summary of the characteristics of each dataset. For each dataset we indicate the total number of samples, the length of each sample, the number of classes and whether the dataset is imbalanced or not.}
\label{table_dataset}
\end{table}
\egroup

\medskip

\noindent
{\bf Methodology.} In order to avoid discrepancies in the accuracy of the method due to the different ranges of values among datasets, input functions have been standardized to have values between 0 and~1. Moreover, when the topological preprocessing is applied, landscapes have been normalized so that the area under each landscape function is equal to~$1$. In doing so, we force the neural network to study the shape of the landscape, rather than only taking into account its actual values. 

The main objective of our study is to compare the ability  of landscape levels to capture information against a baseline of the raw data with the only preprocessing of standardization. Furthermore, to assess that the selected landscape levels are sufficient to classify, the results of feeding a neural network with the full landscape and the results of using only the selected levels are compared.

\begin{figure}[htb]
    \centering
    \includegraphics[width=0.8\textwidth]{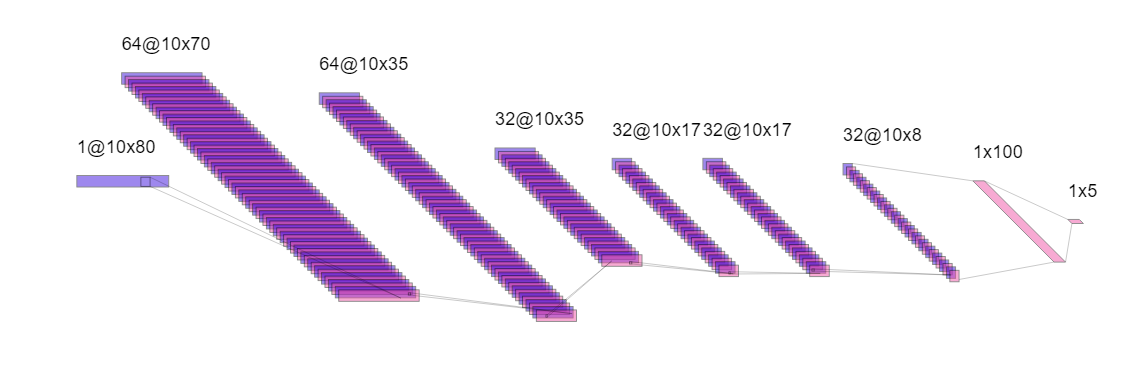}
    \caption{Architecture of the neural network designed for this study. The gating layer is placed immediately before the first dense layer (pink) when landscapes are used as input.}
    \label{fig:architecture}
\end{figure}

The architecture of the neural network is as follows: 
three convolutional layers combined with row-preserving max pooling layers followed by two dense layers 
(Fig.\;\ref{fig:architecture}). Our gating layer is used for selection and attribution purposes and it is only present when landscape levels are used as input. In such case, the gating layer is placed between the last max pooling layer and the first dense layer. The experiments are conducted using a 5-fold cross-validation. 
Training sets amount to 80\% of each dataset.
The neural network is trained during 240 epochs, with a starting learning rate of  $0.01$ that is divided by 5 every 100 epochs. This architecture has been chosen to be rather generic, without attempting to achieve the highest possible accuracy, neither with the original data nor by means of landscapes. Our purpose was to assess the validity our method while avoiding possible particularities due to a tailored choice of an optimal architecture.

As for performance metrics, only accuracy is taken into account in this article. 

\subsection{Validation of the method}
\label{subsection3-1}

\subsubsection{Performance results}
We carried out the same experiment for $9$ different datasets from \cite{UCRdataset} to verify the stability of the results (Table\;\ref{accuracies}). For each dataset, we ran a neural network (Fig.\;\ref{fig:architecture}) with three different inputs: the original data, a sequence of persistence landscape levels, and a selected subset of levels.
Since the length of the full sequence of nonzero landscape levels is variable, we chose the first 10 levels $\lambda_1,\dots,\lambda_{10}$ since in most cases the 10th level was already zero, and fixing a larger number of landscape levels caused memory difficulties during the training process without a significant increase in accuracy. 

Subsequently, the selection of a smaller number of principal landscape levels was made by choosing the highest weights provided by the gating layer. The number of selected levels ranged from $2$ to $5$ depending on the dataset (Fig.\;\ref{fig:error_bars_weights}). Further details about the selection of an appropriate subset of landscape levels are given in \S\;\ref{ranking}.

Table\;\ref{accuracies} shows the average accuracy and standard deviation of each experiment using $5$-fold cross-validation. The table contains average accuracy results using raw data, unnormalized landscapes, normalized landscapes, and a selected subset of normalized landscape levels.
The results show that landscapes achieve sufficiently high classification accuracies, especially when they are normalized (third and fourth columns). In that respect, landscape accuracies are statistically comparable up to one standard deviation to using raw data in four out of the nine datasets. 

In Table\;\ref{accuracies},
the results obtained by TDA-based strategies that are statistically comparable among them ---including the method that achieved maximum accuracy--- are highlighted in bold font. Unnormalized landscapes consistently miss relevant information in most cases, and this is translated into a significant reduction in accuracy. It is also remarkable that the selected landscape levels achieve similar performances as whole (normalized) landscapes. This reinforces the hypothesis that most of the information contained in data is captured by a small subset of landscape levels. 

In the PhalangesOC dataset,
normalized landscapes perform even better than the original data. As pointed out in the Discussion, this could be due to the inherent elastic deformation invariance provided by the landscape representation.

\bgroup
\def\arraystretch{1.3}
\begin{table}[H]
\centering
\begin{tabular}{l|c|cccc}
\textbf{Dataset}
& \textbf{Raw data} 
& \hspace{-0.1cm}\textbf{Unnormalized} & \hspace{-0.1cm}\textbf{Normalized}
& \hspace{0.1cm} \textbf{Selected} & No. \\ \hline
\textbf{ECG5000} 
& $94.72 \pm 0.7$
& ${\bf 92.96 \pm 0.5}$
& ${\bf 93.12 \pm 0.4}$ 
& \hspace{0.1cm} ${\bf 92.88 \pm 0.3}$ & 3 \\
\textbf{FreezerRT} & $99.53 \pm 0.4$
& $63.70 \pm 4.2$
& ${\bf 88.97 \pm 0.3}$
& \hspace{0.1cm} ${\bf 88.70 \pm 1.6}$ & 2 \\
\textbf{HandOutlines}       
& $89.20 \pm 1.7$
& $75.77 \pm 4.6$
& ${\bf 85.26 \pm 2.0}$
& \hspace{0.1cm} $81.90 \pm 2.0$ & 5 \\
\textbf{ItalyPowerD}  
& $97.73 \pm 1.1$
& $87.18 \pm 3.4$
& ${\bf 89.45 \pm 1. 7}$ 
& \hspace{0.1cm} ${\bf 90.36 \pm 1.3}$ & 2 \\
\textbf{MoteStrain}         
& $90.98 \pm 1.1$ 
& $71.84 \pm 1.8$
& ${\bf 77.33 \pm 2.8}$ 
& \hspace{0.1cm} ${\bf 76.31 \pm 1.7}$ & 3 \\
\textbf{PhalangesOC} 
& $64.02 \pm 2.1$
& $63.95 \pm 4.2$ 
& ${\bf 68.95 \pm 0.9}$
& \hspace{0.1cm} ${\bf 69.21 \pm 0.8}$
& 4\\
\textbf{StarlightCurves}
& $95.70 \pm 0.4$ 
& $89.82 \pm 2.3$
& ${\bf 94.92 \pm 0.1}$ 
& \hspace{0.1cm} ${\bf 95.22 \pm 0.5}$ & 3 \\
\textbf{Wafer}              
& $99.65 \pm 0.2$ 
& $90.86 \pm 0.8$
& ${\bf 98.63 \pm 0.3}$
& \hspace{0.1cm} ${\bf 98.79 \pm 0.4}$ & 3 \\
\textbf{Yoga}               
& $82.48 \pm 1.6$
& $64.21 \pm 4.1$
& $75.36 \pm 1.1$ 
& \hspace{0.1cm} ${\bf 78.33 \pm 1.0}$ & 4 \\ 
\hline
\end{tabular}
\caption{Average accuracies (given as percentages) and standard deviations on test sets from five runs of a neural network (Fig.\;\ref{fig:architecture}) for nine signal datasets. Accuracies obtained from original data (first column) are compared with those obtained from the first 10 landscape levels without area normalization (second column) and with area normalization (third column),
and from the most informative landscape levels (fourth colum). The last column indicates how many landscape levels were selected in each case. Statistically comparable accuracies among TDA-based strategies appear in boldface.}
\label{accuracies}
\end{table}
\egroup

\subsubsection{Ranking of landscape levels}
\label{ranking}
The keystone of our process is to be able to identify which landscape levels carry the highest amount of information for classification outcomes.
The gating layer multiplies each landscape level $\lambda_k$ (with $k=1,\dots,10$) by 
a learnable weight $w_k$ with $0\le w_k\le 1$. 
After the full training process of the neural network, the resulting weights are used to attribute importance to each landscape level. 

\begin{figure}[htb]
    \begin{subfigure}{.32\textwidth}
      \centering
      \includegraphics[width=1.1\textwidth]{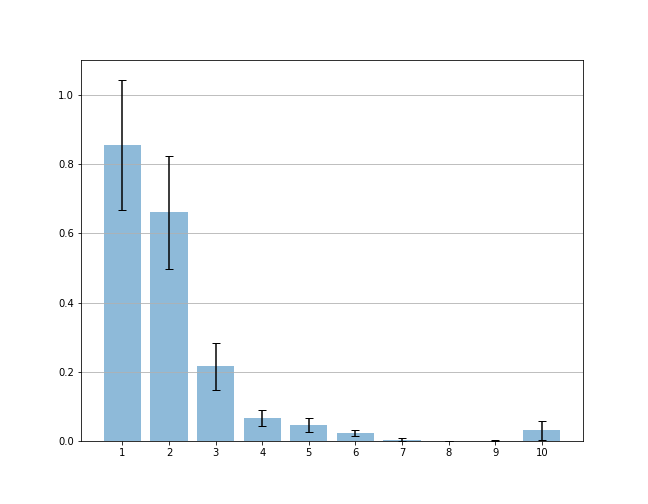}  
      \caption{ECG5000}
    \end{subfigure}\hspace{0.3em}
    \begin{subfigure}{.32\textwidth}
      \centering
      \includegraphics[width=1.1\textwidth]{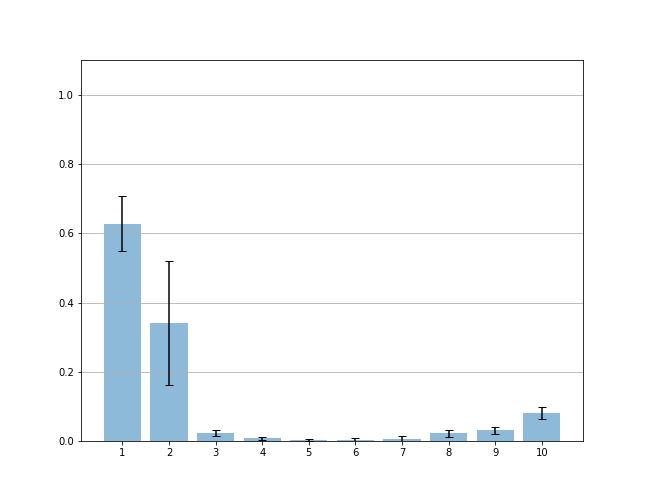}
      \caption{FreezerRegularTrain}
    \end{subfigure}\hspace{0.3em}
        \begin{subfigure}{.32\textwidth}
      \centering
      \includegraphics[width=1.1\textwidth]{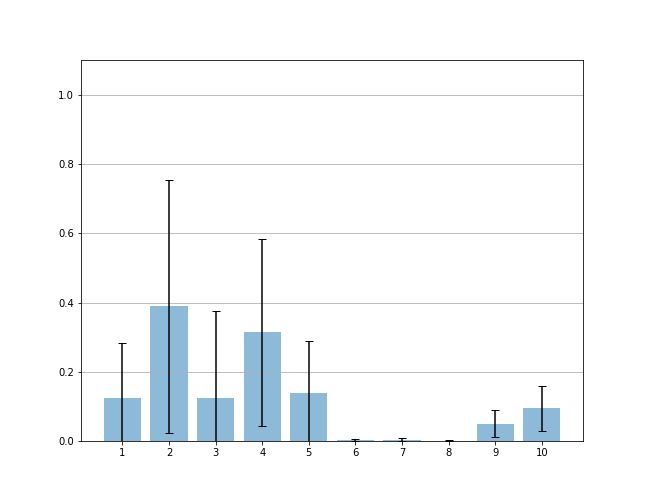}  
      \caption{HandOutlines}
    \end{subfigure}
    
    \begin{subfigure}{.32\textwidth}
      \centering
      \includegraphics[width=1.1\textwidth]{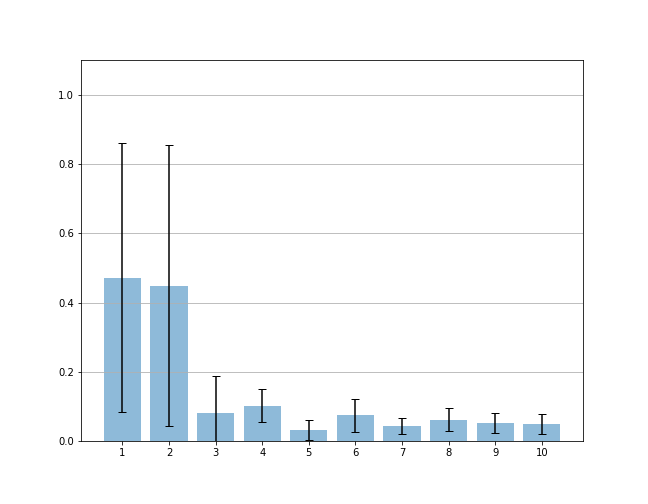}  
      \caption{ItalyPowerDemand}
    \end{subfigure}\hspace{0.3em}
        \begin{subfigure}{.32\textwidth}
      \centering
      \includegraphics[width=1.1\textwidth]{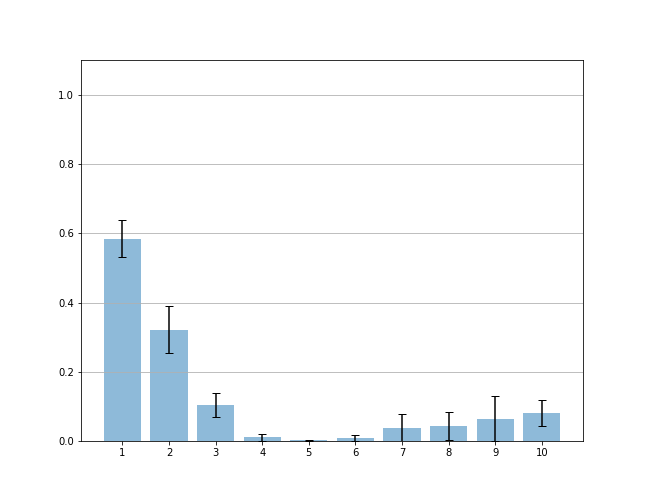}  
      \caption{MoteStrain}
    \end{subfigure}\hspace{0.3em}
        \begin{subfigure}{.32\textwidth}
      \centering
      \includegraphics[width=1.1\textwidth]{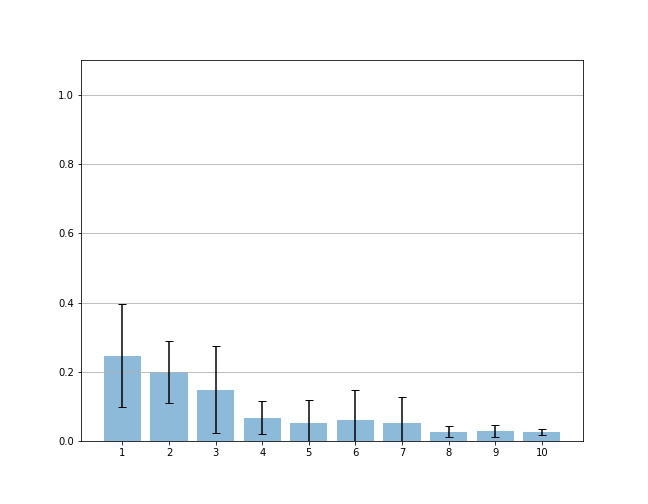}  
      \caption{PhalangesOutlinesCorrect}
    \end{subfigure}
    
    \begin{subfigure}{.32\textwidth}
      \centering
      \includegraphics[width=1.1\textwidth]{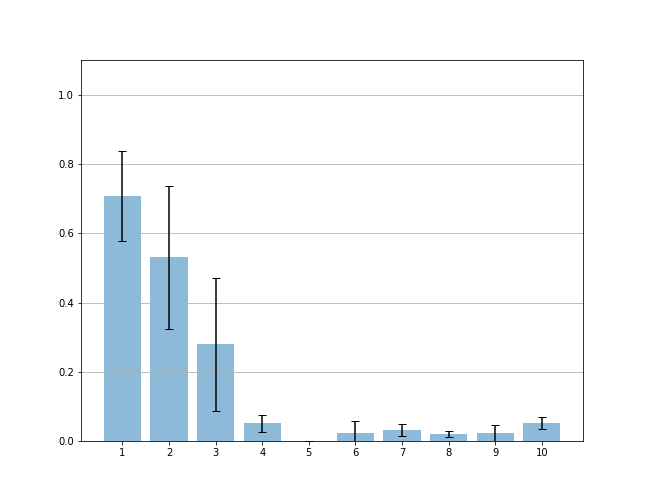}  
      \caption{StarLightCurves}
    \end{subfigure}\hspace{0.3em}
        \begin{subfigure}{.32\textwidth}
      \centering
      \includegraphics[width=1.1\textwidth]{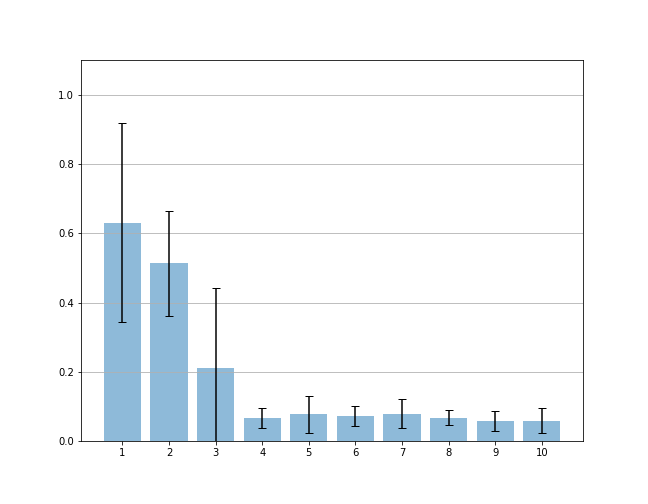}  
      \caption{Wafer}
    \end{subfigure}\hspace{0.3em}
        \begin{subfigure}{.32\textwidth}
      \centering
      \includegraphics[width=1.1\textwidth]{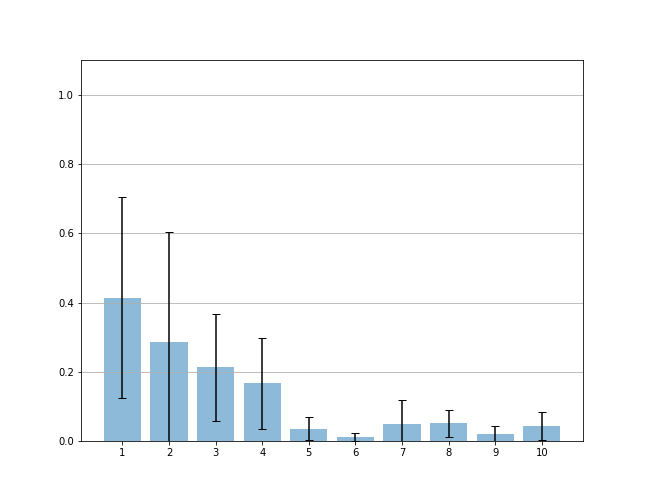}  
      \caption{Yoga}
    \end{subfigure}
    \caption{Average weights and standard deviations of the first ten landscape levels for nine datasets after five runs of a neural network (Fig.\;\ref{fig:architecture}) equipped with a gating layer.}
    \label{fig:error_bars_weights}
\end{figure}

To ensure significance,
we performed the experiment five times and recorded the mean weight value and standard deviation for each landscape level, as seen in Fig.\;\ref{fig:error_bars_weights}. 
Although there is no obvious numerical method to determine the number of landscape levels that should be considered important in view of their weights, we used the following criterion.
If $w_k<\frac12 w_{k-1}$ for some~$k$, we call $k$ a \emph{significant drop}. If $k$ is the largest significant drop with $w_{k-1}>0.1$, then we select $\lambda_1,\dots,\lambda_{k-1}$ as most important landscape levels. If there is no significant drop with $w_{k-1}>0.1$, then we pick the smallest $k$ such that $w_1+\cdots+w_{k-1}>w_k+\cdots+w_{10}$ and also select $\lambda_1,\dots,\lambda_{k-1}$.

With very few exceptions, the network regards the first landscape levels as more important.
These contain information of the most persistent topological features of each signal (connected components of sublevel sets). The first 10 levels were used in all the experiments. 
In some cases ---namely, (d) and (f) in Fig.\;\ref{fig:error_bars_weights}--- landscape levels $\lambda_k$ with $k>6$ were zero for all samples in the dataset. In these cases, the gating layer assigned small but not necessarily zero weights to the null levels.

It is remarkable that the terminal landscape level (i.e., the 10th in our study) tends to be consistently more relevant than the immediately precedent ones, except in those cases where it is zero for the whole dataset. This suggests that the terminal landscape level may convey discriminant information, deserving further study.

\begin{figure}[htb]
    \centering
    \includegraphics[width=.7\textwidth]{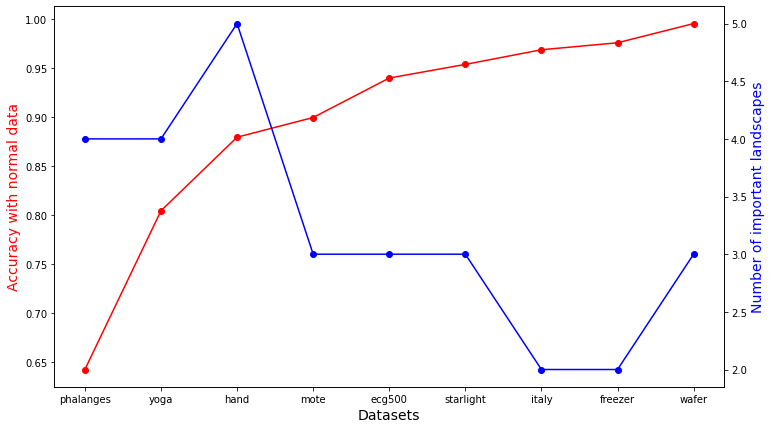}
    \caption{Inverse relationship between the accuracy of our neural network (red) trained with the original raw data and the number of landscape levels (blue) that were selected as important. Datasets in the horizontal axis are ordered by increasing accuracy.}
    \label{fig:relation_weight_accuracy}
\end{figure}

Fig.\;\ref{fig:error_bars_weights} shows that for certain datasets all weights are below $0.4$, specifically (c) and~(f), and marginally also~(i).
Looking at Table\;\ref{accuracies}, we find that these datasets are precisely the ones that 
yield accuracies below 90\% on test sets after the neural network had been trained with the original data.
The datasets where the original data achieved a higher classification accuracy coincide with those with a smallest number of important landscape levels. Indeed, Fig.\;\ref{fig:relation_weight_accuracy} shows an inverse relationship between accuracies and the number of selected landscape levels.

As examples of unfavorable cases, we now discuss results obtained with the datasets FordA and TwoPatterns from \cite{UCRdataset}. 
These datasets share a common property, namely they consist of wave-like signals with a varying wavelength and the key information to classify them is the $x$\nobreakdash-coordinate where the changes in the waves are happening. 
In one of them (FordA), 
the original data are difficult to classify, while in the other one (TwoPatterns) the original data are easily classifiable. In both cases, replacing the data by persistence landscapes erases the relevant information for a neural network classifier ---since landscapes are invariant under wavelength changes if amplitude is preserved---
and thus we obtain low accuracy and considerable overfitting if landscapes are used instead of raw data.

\begin{figure}[htb]
    \centering
    \begin{subfigure}{.38\textwidth}
      \centering
      \includegraphics[width=\textwidth]{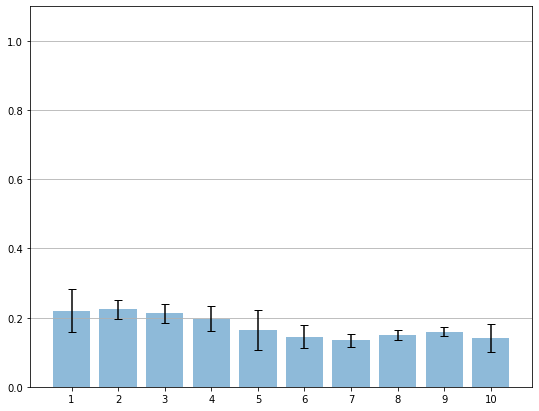}  
      \caption{FordA}
    \end{subfigure}\hspace{3em}
    \begin{subfigure}{.38\textwidth}
      \centering
      \includegraphics[width=\textwidth]{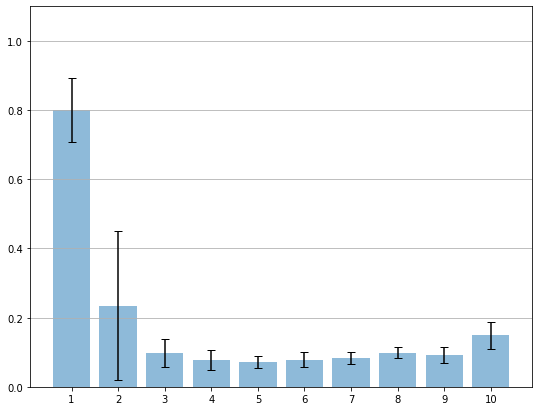} 
      \caption{TwoPatterns}
    \end{subfigure}
    
    \caption{Average weights and standard deviations of the first ten landscape levels for two datasets after five runs of a neural network (Fig.\;\ref{fig:architecture}) equipped with a gating layer.}
    \label{fig:weight_importance_2}
\end{figure}

In Fig.\;\ref{fig:weight_importance_2} we see that,
for the FordA datasets (where the neural network has trouble classifying even with the original data) the weights of persistence landscape levels are all similar and with a low relevance. In contrast, in the TwoPatterns case we see a clear ranking of the first landscape levels. 
Hence, landscape selection yields meaningful information about the dataset even in disadvantageous situations,
since there is a consistent inverse relationship between the ability of the neural network to correctly classify the original data and the number of important landscape levels found through our method. 
In conclusion,
Fig.\;\ref{fig:relation_weight_accuracy} and Fig.\;\ref{fig:weight_importance_2} 
provide evidence that the outcome of landscape level selection can be related to how well a neural network can perform.

\subsection{A use case: Results of a heartbeat analysis}
\label{subsection3-2}

As an application case, we used our algorithm for a classification of electrocardiogram signals (ECG) from the MIT-BIH Arrhytmia Database \cite{MoodyMark2001} for evaluation of arrhytmia detectors.
The dataset can be retrieved from~\cite{kaggle} and it includes 48 half-hour excerpts of 24-hour ECG recordings obtained from 47 subjects (25 men aged 32 to 89 years and 22 women aged 23 to 89 years) studied between 1975 and 1979.
Our data sample includes 87,554 heartbeats of five classes: one corresponding to normal beats (82.77\%); three classes corresponding to different arrhythmia types, namely supraventricular premature beats (2.54\%), premature ventricular contraction (6.61\%), and fusion of ventricular and normal beats (0.73\%); and one class for unidentifiable heartbeats (7.35\%).

\begin{figure}[H]
    \centering
    \includegraphics[width=.5\textwidth]{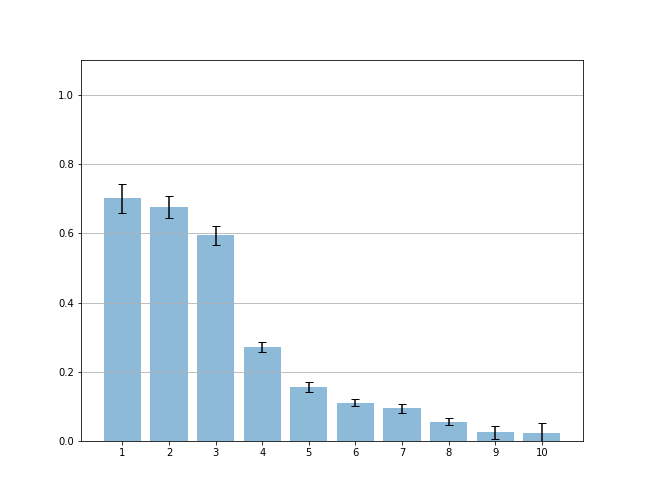}
    \vspace*{-0.5cm}
    \caption{Average weights and standard deviations of the first ten landscape levels for a sample of the MIT-BIH Arrhytmia Database after five runs of a neural network (Fig.\;\ref{fig:architecture}).}
    \label{fig:weights}
\end{figure}

Table~\ref{accuracy} shows average accuracy 
after a 5-fold cross-validation. The classification accuracy of our neural network (Fig.\;\ref{fig:architecture}) fed with the original unprocessed signal (98.41\%) is compared with the accuracy of the same architecture using a 10-level landscape (94.55\%) and using only the three most important landscape levels (94.00\%). Landscapes were area-normalized since Table~\ref{accuracies} evidenced an advantage of normalized landscapes versus unnormalized ones.
The choice of three levels was based on weights assigned by the network, as shown in Fig.\;\ref{fig:weights}, where $k=4$ is the largest significant drop.

\begin{figure}[htb]
\centering
    \begin{subfigure}{.45\textwidth}
      \centering
      \includegraphics[width=1\textwidth]{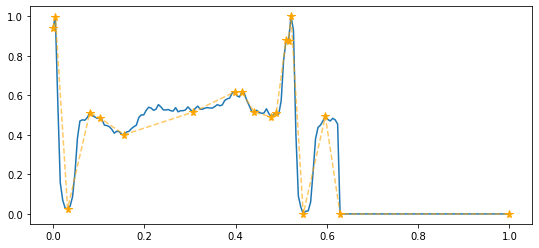}
    \caption{Normal heartbeat}
    \label{fig:normal}
    \end{subfigure}\hspace{1em}
    \begin{subfigure}{.45\textwidth}
      \centering
      \includegraphics[width=\textwidth]{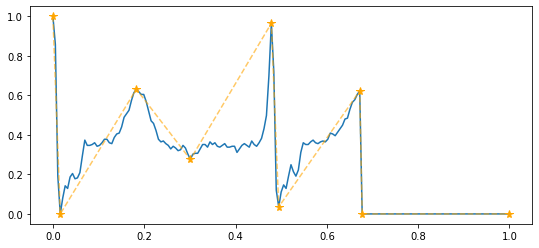}
    \caption{Supraventricular premature beat}
    \label{fig:SPB}
    \end{subfigure}
    
    \begin{subfigure}{.45\textwidth}
      \centering
      \includegraphics[width=\textwidth]{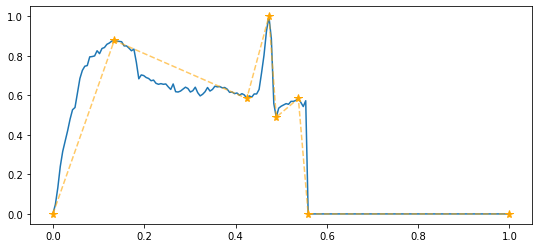}
    \caption{Premature ventricular contraction}
    \label{fig:PVC1}
    \end{subfigure}\hspace{1em}
    \begin{subfigure}{.45\textwidth}
      \centering
      \includegraphics[width=\textwidth]{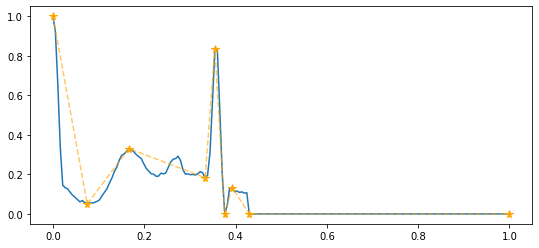}
    \caption{Fusion of ventricular and normal beat}
    \label{fig:PVC2}
    \end{subfigure}
    
    \begin{subfigure}{.45\textwidth}
      \centering
      \includegraphics[width=\textwidth]{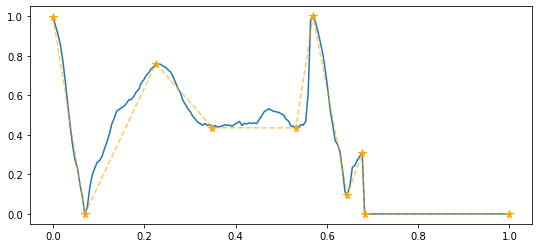}
    \caption{Unclassifiable beat}
    \end{subfigure}
    
    \caption{Partial reconstruction of ECG graphs using the three most important landscape levels for each of four types of heartbeats.}
\label{fig:reconstruction}
\end{figure}

Next we used the partial reconstruction technique described in detail in  \cite[Section~3]{ACO1} in four examples, corresponding to the classes of (a) normal heartbeats, (b) supraventricular premature beats, (c) premature ventricular contraction, and (d) fusion of ventricular and normal beats.
Three landscape levels were used for approximation in each case. The outcome is shown in Fig.\;\ref{fig:reconstruction}.

\begin{figure}[htb]
    \centering
    \includegraphics[width = 0.75 \textwidth]{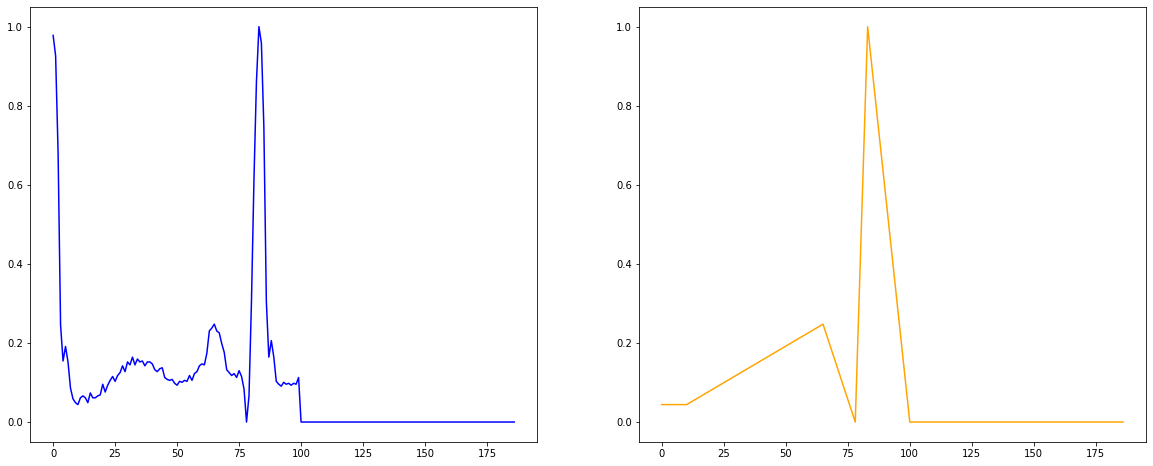}
    \caption{An ECG signal function (left) and its approximate reconstruction from a set of selected landscape levels (right).}
    \label{fig:reconone}
\end{figure}

Each landscape function $\lambda_k$ was paired with a list of $y$-values of critical points of the given signal $f$ as specified in \cite[Proposition~3.1]{ACO1}. Hence we obtained a list of $y$-values of critical points of $f$ associated with the subset of selected landscape levels.
The values in this list were compared with the list of all critical points of $f$ in order to obtain the matching $x$-values, and a new graph was drawn by joining the resulting critical points of $f$ in the order of their $x$-coordinates, as in Fig.\;\ref{fig:reconone}.
The procedure is detailed below in Algorithms~1, 2 and~3.
The resulting simplified graphs (Fig.\;\ref{fig:reconstruction}) mark the points of interest, according to the neural network used in our experiment, for the classification of ECG samples. Thus they encode the most relevant information on which the network focused for its task.

\begin{table}[htb]
\centering
\begin{tabular}{lcccc}
& \textbf{Raw data} & \;\textbf{10 levels} & \textbf{3 levels} & \textbf{Reconstructed} \\ \hline \\[-0.25cm]
\textbf{Accuracy} \;\; & $98.41 \pm 0.09$ & \;$94.55 \pm 0.16$     & $94.00 \pm 0.13$ & $97.04 \pm 0.14$ \\[0.1cm] \hline
\end{tabular}
\caption{Accuracy of classification (given in percentages) of our neural network (Fig.\;\ref{fig:architecture}) fed with unprocessed data versus processed data with ten landscape levels and processed data using the most significant three landscape levels, as well as data approximately reconstructed by means of three landscape levels.}
\label{accuracy}
\end{table}

We subsequently introduced the simplified reconstructions of the wave functions (Fig.\;\ref{fig:reconone}) into the network in order to check if the data features distilled by our reconstruction method were sufficient for the network's classifications task. The results can be seen in Table~\ref{accuracy} and indicate that the simplified signals gave rise to similar accuracies (97.04\%) as the original data (98.41\%). 

\subsection{Invariance under translations}
\label{subsection3-3}

Persistence summaries are not altered by horizontal shifts of signals and hence the accuracy of a classification task based on landscapes is invariant under such shifts.
However, shifts may cause a loss of classification accuracy by a neural network fed with the original data. To demonstrate this effect, we used the same ECG dataset from \S\,\ref{subsection3-2}, yet we modified each heartbeat by adding a  number of zeros randomly split between the beginning and the end of the beat signal. 
Thus, while in the original dataset each heartbeat was represented by a vector of length 187, in our experiment we introduced zeros so that the length was increased to~374. 

\begin{table}[H]
\centering
\begin{tabular}{lcc}
& \textbf{Raw data} & \;\textbf{Double length}  
\\ \hline \\[-0.25cm]
\textbf{Accuracy} \;\; & $98.41 \pm 0.09$ & \;$96.77 \pm 0.10$
\\[0.1cm] \hline
\end{tabular}
\caption{Accuracies (given in percentages) of our neural network fed with unmodified data versus modified data by inserting zero segments at the beginning and end of each signal so as to duplicate the length of the signals (second column).}
\label{tres}
\end{table}

Classification of the shifted ECG graphs by means of the same neural network as in \S\,\ref{subsection3-2} with five repetitions resulted in lower accuracy (Table~\ref{tres}) than with the original data. However, shifts do not alter the evolution of connected components of sublevel sets of the graphs and therefore the landscapes associated with the shifted graphs are the same as those of the original data. This illustrates that the use of persistence descriptors can be advantageous in practical cases where translations of data signals are unsubstantial in nature and nevertheless can be misinterpreted by a neural network.

\vspace*{-0.1cm}

\section{Discussion}
\label{discussion}

\vspace*{-0.1cm}

This article highlights an instance of the usefulness of topological data analysis in machine learning, specifically in the field of interpretability of outcomes of neural networks. Our procedure enabled us to distill partial information from the given data sufficiently relevant for classification purposes without a significant loss of accuracy. We used landscapes as descriptors of persistence of sublevel sets of signals, since landscapes come with a hierarchy of levels that enables us to rank the importance of each level by means of weights assigned by a gating layer in a neural network.

We confirmed that using the whole persistence landscape is not necessary for an accurate classification of signals: once we have identified the subset of landscape levels that is most important for the network, running the experiment with only this subset of levels yields a statistically comparable accuracy (Table\;\ref{accuracies}).

A novelty of this study is normalization of landscape functions so that the area below the graph is constantly equal to one. This was conceived as an attempt to feed the neural network with \emph{shapes} rather than \emph{magnitudes}. As Table\;\ref{accuracies} shows, the accuracies obtained with normalized landscapes were higher than those obtained prior to normalization.
Furthermore, the standard deviation of accuracy is lower after normalization in most cases, suggesting that normalization enhances stability. 

In addition, our method allows us to partially reconstruct the given signals using the set of selected landscape levels, thus depicting which features of the data are most relevant for classification by means of the chosen architecture.
Persistence descriptors are not injective in general and cannot be used to recover the data except in some cases where a collection of directional persistence diagrams are considered \cite{beltonetal, ChazalOudot2008,ACO1,turneretal}. However, our problem at hand is different: we need not fully 
reconstruct a function with the only knowledge of 
its persistence diagram,
but rather attribute the persistence descriptors to the original information. Hence, our reconstruction task consists of matching points in the persistence landscape with corresponding parts of the given signals.
Indeed, $y$-values of critical points of signals 
are determined by points in 
persistence diagrams as in \cite[Section~1]{ACO1}. 

Since we are discretizing the given signals, difficulties regarding numerical precision may arise. When we are searching for peaks in landscape functions, the corresponding $y$-values of critical points in signals can be computed up to the chosen precision. When comparing those $y$-values with the original functions, we cannot expect a zero difference, since we are dealing with approximations. 
Instead, we have to take values that are below a certain threshold. 
In order to avoid picking $x$-values that do not correspond to the correct critical points, each time we had a candidate $x$-value we checked in the original function that it came indeed from a minimum or a maximum. 

A feature of our method is that persistence diagrams of sublevel sets of signals do not capture information about the distribution of data along the $x$-axis, but only along the $y$-axis. As a consequence, 
persistence summaries such as landscapes are invariant with respect to translations or scale changes on the $x$-axis, or, more generally, with respect to horizontal elastic deformations.
This can be a disadvantage for the use of persistent homology in cases when, for example, the wavelength of periodic or almost periodic functions is crucial for classification purposes, as illustrated 
by the datasets FordA and TwoPatterns in Subsection~\ref{ranking}.
However, it can be an advantage if expansion or contraction along the $x$-axis produces undesired effects, as in the case of bradycardia and tachycardia in \cite{meryll2019betti} or in the experiment made in Subsection~\ref{subsection3-3}.
Hence, the use of persistence summaries can serve to remove spuriousness due to shifts on the $x$-coordinate without a physical significance; e.g., when time series are segmented into shorter signals or when random horizontal segments occur within signals.

\enlargethispage{0.5cm}

Regardless of the effect on performance metrics of using persistence descriptors  instead of raw data, by using landscapes we gain insight about key patterns used to classify the given data, which makes the process more trustworthy.
Thus, our method not only provides information about the focus of the network's learning process but it also serves to explore and better understand the dataset.

\printbibliography

\newpage

{\begin{algorithm}[H]
\SetKwData{crits}{crits}
\SetKwInOut{Input}{Input}
\SetKwInOut{Output}{output}

\KwIn{$l$ vector of landscape data, $t$ vector of $t$-coordinates of landscape data\;}
\KwOut{List of $y$-values of critical points\;}
\BlankLine
$\crits \leftarrow \{ \}$\;
\For{$i\gets0$ \KwTo $|l|$}{
    \uIf{$l_i$ is a take-off vertex}{
         $\crits \leftarrow \crits \cup \{t_i\}$\;
    }\uElseIf{$l_i$ is a local minimum or a local maximum}{
        $\crits \leftarrow \crits \cup \{t_i\}$\;
        $\crits_i \leftarrow 2(\crits_i - 0.5\,\crits_{i-1})$\;}
    }
\KwRet{$\crits$}
\caption{Get $y$-values of critical points}
\end{algorithm}
}

\vspace*{1cm}

{\begin{algorithm}[H]
\SetKwData{crits}{crits}
\SetKwData{xcrits}{xcrits}
\SetKwData{candcrits}{candcrits}

\SetKwInOut{Input}{Input}
\SetKwInOut{Output}{output}

\KwIn{$\crits$ list of $y$-values of critical points, $original$ vector of original data, $x$ vector of $x$-coordinates of original data\;}
\KwOut{List of $x$-values of critical points\;}
\BlankLine
$\candcrits \leftarrow \{ \}$\;
\For{$crit \in \crits$}{
    $compare \leftarrow (original - crit)^2$\;
    \For{$i\gets0$ \KwTo $|compare|$}{
        \uIf{$compare_i < 10^{-4}$}{
        $\candcrits \leftarrow \candcrits \cup \{i\}$\;
        }
    }
}

$\xcrits \leftarrow \{ \}$\;
\For{$i \in \candcrits$}{
    \uIf{$original_i$ is a local minimum or a local maximum}{
        $\xcrits \leftarrow \xcrits \cup \{x_i\}$\;
        }
}
\KwRet{$\xcrits$}
\caption{Get $x$-values of critical points}
\end{algorithm}
}

\vspace*{1cm}

{\begin{algorithm}[H]
\SetKwData{crits}{crits}
\SetKwData{landscapes}{landscapes}
\SetKwData{original}{original}

\SetKwInOut{Input}{Input}
\SetKwInOut{Output}{output}

\KwIn{$\landscapes$ a list of landscape levels, $original$ vector of original data\;}
\KwOut{List of selected critical points of the given function\;}
\BlankLine
$\crits \leftarrow \{ \}$\;
\For{$land \in \landscapes$}{
    $ candcrits \leftarrow \{ $\texttt{get\_y\_values}$(land) \}$\;
    $\crits \leftarrow \crits \cup \{$\texttt{get\_x\_values}$(candcrits, original) \} $\;
    }
$\crits \leftarrow $\texttt{sort}$(\crits)$\;
\KwRet{$\crits$}
\caption{Subset of critical points of a function associated with a set of landscape levels}
\label{aquest}
\end{algorithm}
}

\end{document}